\documentclass[runningheads]{llncs}

\usepackage{amssymb,amsmath,amsfonts}
\usepackage{graphicx}
\usepackage{url}
\usepackage{color}
\usepackage{amsfonts}
\usepackage{blindtext}
\usepackage{mathrsfs}
\usepackage{bbm}
\usepackage{url}
\usepackage{paralist}
\usepackage[english]{babel}
\usepackage{booktabs}
\usepackage{multicol}
\usepackage{multirow}
\usepackage{subfig}
\usepackage{amsmath}
\usepackage{pdfpages}
\usepackage{threeparttable}
\usepackage{float}
\usepackage{placeins}
\begin{document}
\title{Automatic Task Requirements Writing Evaluation via Machine Reading Comprehension}
\titlerunning{Automatic Task Requirements Writing Evaluation}
\author{Shiting Xu, Guowei Xu, Peilei Jia, Wenbiao Ding\thanks{Corresponding Author: Wenbiao Ding}, Zhongqin Wu, Zitao Liu}
% %

\authorrunning{S. Xu et al.}
\institute{TAL Education Group, Beijing, China \\
\email{\{xushiting, xuguowei, jiapeilei, dingwenbiao, wuzhongqin, liuzitao\}@tal.com}}
\maketitle              % typeset the header of the contribution
\begin{abstract}

Task requirements (TRs) writing is an important question type in Key English Test and Preliminary English Test. A TR writing question may include multiple requirements and a high-quality essay must respond to each requirement thoroughly and accurately. However, the limited teacher resources prevent students from getting detailed grading instantly. The majority of existing automatic essay scoring systems focus on giving a holistic score but rarely provide reasons to support it. In this paper, we proposed an end-to-end framework based on machine reading comprehension (MRC) to address this problem to some extent. The framework not only detects whether an essay responds to a requirement question, but clearly marks where the essay answers the question. Our framework consists of three modules: question normalization module, ELECTRA based MRC module and response locating module. We extensively explore state-of-the-art MRC methods. Our approach achieves 0.93 accuracy score and 0.85 F1 score on a real-world educational dataset. To encourage reproducible results, we make our code publicly available at \url{https://github.com/aied2021TRMRC/AIED_2021_TRMRC_code}.

\end{abstract}

\section{Introduction}
\label{sec:intro}

Key English Test\footnote{https://www.cambridgeenglish.org/exams-and-tests/key} (KET) and Preliminary English Test\footnote{https://www.cambridgeenglish.org/exams-and-tests/preliminary} (PET) are examinations to assess the communication ability of the test taker in practical situations. In PET and KET, there are a variety of question types, including speaking, reading, listening, and writing. In writing questions, examinees are not only required to write an essay precisely and correctly but need to make responses to the \textbf{Task Requirements (TRs)}. According to official scoring instructions, an essay with poor task achievements should be assigned a low grade. Some examples of TR writing questions are shown in Table \ref{table_task_examples}.

\begin{table}[H]
\centering
\caption{\label{table_task_examples}Examples of TR writing questions in PET.}
\begin{threeparttable}
\begin{tabular}{ll}
%  & \multicolumn{1}{c}{Task Requirements Writing Questions}  
\\[-1.8ex]                                                                                                                                                                                                                                                                                
\hline \\[-1.8ex]
1 & \textit{\begin{tabular}[c]{@{}l@{}}A TV company came to your school yesterday to make film.\\ \\[-2.5ex] Write an email to your English friend Alice. In your email, you should\\ \\[-2 ex] \textbf{* explain why the TV company chose your school} \\ \textbf{* tell her who or what they filmed} \\ \textbf{* say when the programme will be shown on television.} \end{tabular}}  \\ \\[-1.8ex] \hline \\[-1.8ex]
2 & \textit{\begin{tabular}[c]{@{}l@{}}You arranged to meet your English friend Sally next Tuesday, \\ but you have to change the time.\\ \\[-2.5ex] Write an email to Sally. In your email, you should \\ \\[-2 ex] \textbf{* suggest a new time to meet on Tuesday}\\ \textbf{* explain why you need to change the time} \\ \textbf{* remind Sally where you arranged to meet}\end{tabular}} \\ \\[-1.8ex] \hline 
\end{tabular}
\begin{tablenotes}\centering
\item[1] Lines begin with $\textbf{*}$ are task requirement questions.	
\end{tablenotes}
\end{threeparttable}
\end{table}

Timely and accurate evaluation on the performance of test-takers, especially informing them of TR achievements of their essays, is essential to improve their writing and communication skills. Such evaluation usually takes experienced teachers a large amount of time as each essay needs to be graded carefully. However, due to the limitation of teacher resources, most English learners cannot get timely assessments on the quality of their essays. Although many researchers studied how to automatically score an essay, most of the current approaches can only provide total scores without enriched supports \cite{taghipour2016neural,dong2017attention,wang2018automatic}. This is not really helpful for students to improve their writing skills.

In natural language processing field, machine reading comprehension (MRC) has been studied for a long time and can be employed to provide details in terms of how well TRs have been achieved in students' essays. In MRC field, the second version of Stanford Question Answering Dataset (SQuAD 2.0) is the most widely used benchmark dataset to evaluate model performance \cite{rajpurkar2018know}. However, our experiments prove that even a model that achieves the best performance on SQuAD 2.0 cannot be directly used on educational scenarios, as there is a significant performance degradation. The main reason is that SQuAD 2.0 is a general-purpose open-source dataset, but there is a huge difference between educational and general-purpose corpora. 

To alleviate these problems, we construct a real-world educational dataset and propose an end-to-end framework based on MRC approach, which uses ELECTRA as a backbone, to detect whether students respond to TRs in their essays \cite{clark2020electra}. Our framework can clearly and accurately locate sentences in student essays that respond to the requirements. Experiments on an educational dataset show that the proposed framework achieves 0.93 accuracy score and 0.85 F1 score, outperforming many existing approaches. We believe that this research can help automatic essay scoring system provide interpretable grading results, thereby helping students improve their writing skills. 

%This provides great convenience for English learners to scrutinize and revise their writing content to meet the TRs which can strengthen their awareness of target task response, and help them improve their writing and communication abilities effectively.

\section{Related Work}
\label{sec:related}
\subsection{Automated Writing Evaluation}
Automated writing evaluation (AWE) has been studied for a long time in both industry and academia \cite{klebanov2020automated,page1966imminence,ke2019automated,wang2020learning}. Since Page and Ellis B published their works in 1996, plenty of automated scoring products and applications, e.g., E-rater, have emerged. Based on AWE, lots of works on automatic essay scoring (AES) have been published \cite{page1966imminence,taghipour2016neural,dong2017attention,wang2018automatic}. However, these works mainly focused on giving a holistic score, which measures the overall quality of an essay. Taghipour explored several neural network models for AWE and outperformed strong baselines without requiring any feature engineering \cite{taghipour2016neural}. Dong proposed a reinforcement learning framework that incorporates quadratic weighted kappa as guidance to optimize the scoring system \cite{dong2017attention}. In recent years, a variety of researches focused on fine-grained essay evaluation \cite{carlile2018give,ke2019give,persing2014modeling}. In Persing's work, they presented a feature-rich approach to score prompt adherence of essays \cite{persing2014modeling}. In Ke's work, they not only predicted a score of thesis strength but also provided more reasons \cite{ke2019give}. Nevertheless, none of these works address the problem of detecting TR achievements in AES systems.

\subsection{Machine Reading Comprehension}
At document level, finding students' response to a TR is similar to extractive and abstractive MRC task in which given several reading materials, the model is expected to answer related questions based on the materials. The MRC models are expected to understand both the context and the question and be able to perform reasoning. In TR writing, we could regard student's essays as reading materials, and the model is supposed to find answers to TRs. If no answer is found, it indicates that the essay does not respond to the requirement.

The early trend of MRC used long short-term memory or convolutional neural network as an encoder of questions and contexts and blended a variety of attention mechanisms, e.g., attention sum, gated attention \cite{hochreiter1997long,o2015introduction,kadlec2016text,dhingra2016gated}. Approaches mimicking the process of how humans do reading comprehension were also proposed, such as multi-hop reasoning \cite{shen2017reasonet,liu2017stochastic,shen2017empirical}.  Recently, pre-trained language models, e.g., BERT, RoBERTa, ALBERT, BART, ELECTRA, became prevalent encoder architectures in MRC and achieved state-of-the-art performance \cite{devlin2018bert,liu2019roberta,lan2019albert,lewis2019bart,clark2020electra}. Besides these improvements and optimizations on the encoder module, research about the decoder in the MRC model also starts to draw attention. Zhang al et. proposed an answer verification method and achieved state-of-the-art single model performance on SQUAD 2.0 benchmark with ELECTRA encoder module \cite{zhang2020retrospective,rajpurkar2018know,rajpurkar2016squad}. 

Another line of research on MRC is how to construct high-quality datasets and lots of works have been done \cite{nguyen2016human,hewlett2016wikireading,trischler2016newsqa,joshi2017triviaqa,rajpurkar2016squad}. Among them, SQuAD is one of the most widely-used reading comprehension benchmarks \cite{rajpurkar2016squad}. However, Rajpurkar et al. showed that the success on SQuAD does not ensure robustness to distracting sentences \cite{rajpurkar2018know,jia2017adversarial}. One reason is that SQuAD focuses on questions for which a correct answer is guaranteed to exist in the context document. Therefore, models only need to select the span that seems most related to the question, instead of checking that the answer is actually entailed by the text. Based on SQuAD, Rajpurkar et al. proposed SQuAD 2.0. To do well on SQuAD 2.0, systems must not only answer questions when possible but determine when no answer is supported by the paragraph and abstain from answering \cite{rajpurkar2018know}.

Comparing with previous AWE works, to the best of our knowledge, we are the first to use a MRC approach to detect TR achievements in educational domain. We also construct a Student Essay Dataset (SED) which can be deemed as SQuAD 2.0 in the educational field and we explore the usage of a combination of SQuAD 2.0 and SED.

\section{Problem Statement}
\label{sec:problem}
%\vspace{-0.5em}
In the TRs writing evaluation task, let $Q$ denote a collection of task requirement questions and $q$ denote a single question in $Q$. Let $t_{q}^i$ denotes the $i$-th token in the question $q$ such that $q=(t_q^1, t_{q}^2, t_{q}^3, \cdots, t_{q}^m)$. $E=(t_e^1, t_e^2, t_e^3, \cdots, t_e^n)$ is an essay written by a student where $t_e^j$ denotes the $j$-th token in the essay $E$. Then the problem is defined as for each requirement $q$, is there a sequential text span $S=(t_e^j, t_e^{j+1}, \cdots, t_e^{j+s})$ in $E$ that responds to the requirement $q$? If such span $S$ exists,  $q$ is achieved and $S$ needs to be extracted from the essay $E$, if not, $q$ is not achieved by $E$.

\section{Method}
\label{sec:method}

\subsection{The Overall Workflow}
\begin{figure}[H]
\centering
\includegraphics[scale=0.24]{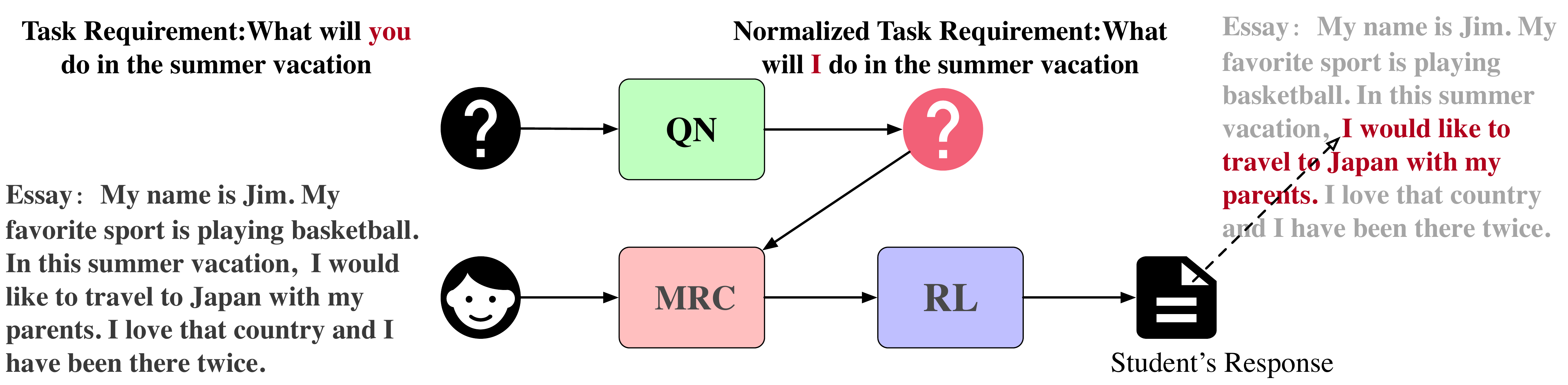}
\caption{The workflow of TRs evaluation framework.}
\label{fig_framework}
\end{figure}

 The overview of our proposed framework is displayed in Figure \ref{fig_framework}. Our approach is mainly composed of three principal modules, question normalization (QN) module, MRC module, and response locating (RL) module.

\subsection{Question Normalization Module}
Task requirement questions are proposed from the perspective of examiners, but essays are from examinees' perspective. This perspective gap brings difficulties to the MRC model. To eliminate the difference, we normalize texts of task requirements with two rule-based methods: switching personal pronouns and deleting redundant words.

\subsubsection{Switch Personal Pronouns} 
We use pre-defined rules to replace personal pronouns in the sentence. For example, a question \textit{``What will you do in the summer vacation ?''} may receive a student's answer \textit{``I will travel to Japan''}. If we change personal pronouns ``you'' in the question, it will be normalized as \textit{``What will I do in the summer vacation ?''}. The normalized question will decrease the difficulties of this task for the models. 

\subsubsection{Delete Redundant Words} 
We define question words such as \textit{``what''}, \textit{``how''}, etc., and then delete redundant words that appear before them. One example of deleting unnecessary words is that we omit the word \textit{``explain''} in the question \textit{``explain why you need to change the time''} and change it to \textit{``why you need to change the time''}. Another instance is that we delete the words \textit{``remind Sally''} in \textit{``remind Sally where you arranged to meet''} and acquire the normalized question \textit{``Where I arranged to meet''}.

\subsection{Machine Reading Comprehension Module}

In MRC module, normalized task requirement question $q$ and the whole essay $E$ are concatenated with a special symbol $[SEP]$. The entire input sequence to MRC model can be described as $T=([CLS],t_{q}^{1},t_{q}^2,t_{q}^3, \cdots, t_{q}^i, \cdots,t_{q}^m,$ $[SEP],$ $t_e^1, t_e^2, t_e^3, \cdots, t_e^j,\cdots, t_e^n)$, where the full length of $T$ is $\tau=m+n+2$.

\subsubsection{ELECTRA Encoder}We use the discriminator module of ELECTRA to encode each token in $T$ into a dense vector. The max length of $T$ is 512 and tokens exceeding the max length will be truncated at the end. We use $h_u^L$ to represent the final layer outputs of ELECTRA at position $u$ which corresponds to the $u$-th token in $T$. We use $H^L = ( h^L_1, . . . , h^L_{\tau})$ to denote the last-layer hidden states of the input sequence, where $H^{L}\in \mathbb{R}^{{\tau}\times{768}}$. ELECTRA model is based on a multi-layer bidirectional Transformer encoder, and multi-head attention network \cite{vaswani2017attention}. Therefore,  $h_u^L$ is able to capture the context of the $u$-th token from  ${q}$ and $E$. The attention function in ELECTRA and the output of layer $l$ are showed in eq.(\ref{eq_attention}). In layer $l$, inputs $Q,K,V$ are computed by $H^{l-1}W_q$ , $H^{l-1}W_k$ , $H^{l-1}W_v$ respectively, where $H^{l-1}$ denotes the output of the previous layer and $W_{q}\in \mathbb{R}^{768\times{d_k}}$, $W_{k}\in \mathbb{R}^{768\times{d_k}}$,  $W_{v}\in \mathbb{R}^{768\times{d_k}}$. Thus $Q,K,V$ have the same dimensions ${\mathbb{R}^{\tau\times{d_k}}}$ where $d_k$ is the dimension of vectors in $K$. 

\begin{equation} 
\begin{split}
Attention(Q,K,V) = softmax(\frac{Q^TK}{\sqrt{d_k}})V \\
H^l = max(0,Attention(Q,K,V)W_1+b_1)W_2+b_2 
\label{eq_attention}
\end{split}
\end{equation}

\begin{figure}\centering
\includegraphics[scale=0.25]{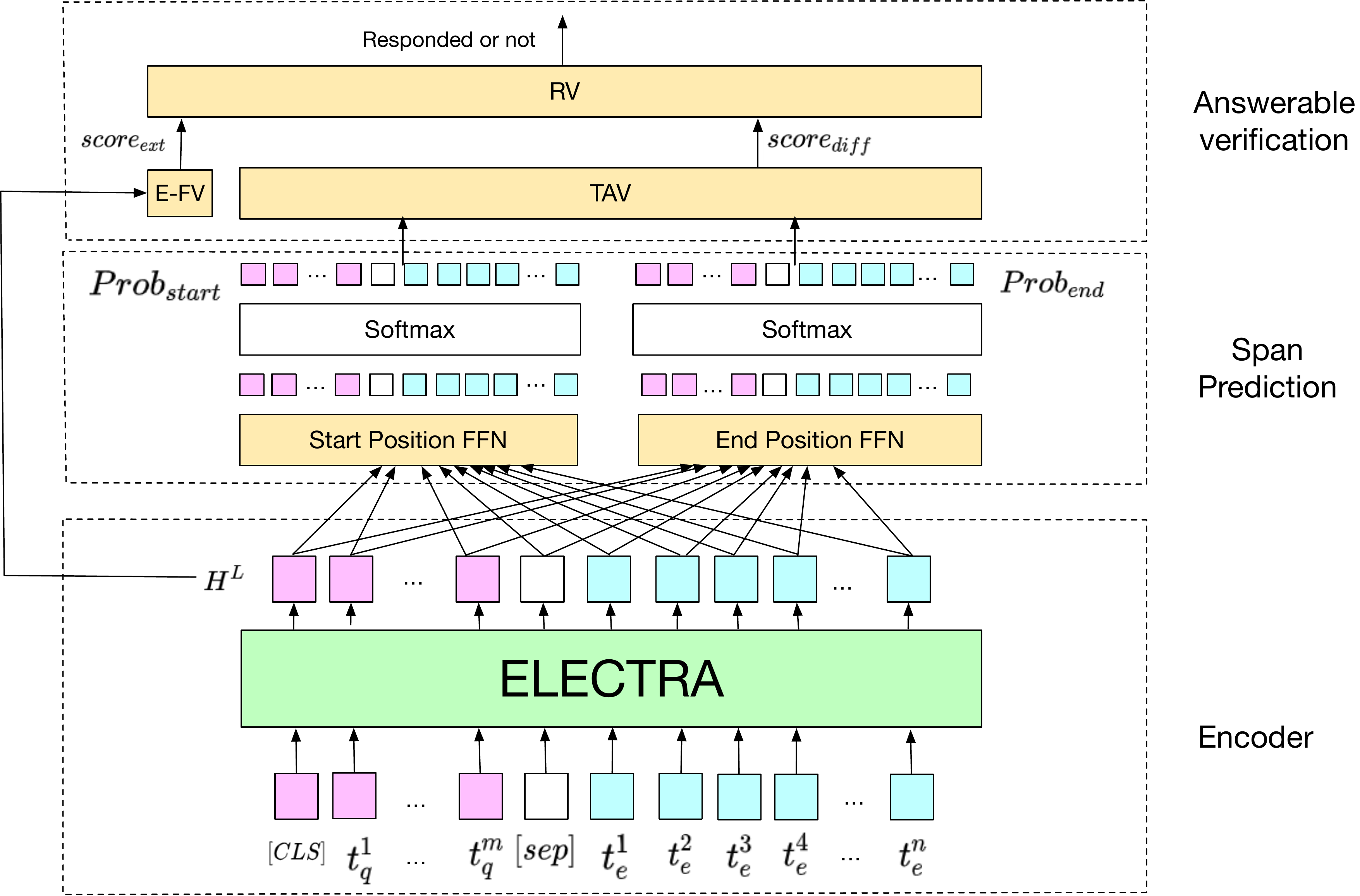}
\caption{The architecture of MRC model}
\label{fig_prlmrc}
\end{figure}

\subsubsection{Span Prediction}
We employ a fully connected layer with softmax operation which takes $H_L$ as input and outputs start and end probabilities of each token in $T$, as shown in eq.(\ref{eq_prob_start}). Let $p^i_{start}$ and $p^i_{end}$ represent the start and end probabilities of $i$-th token in $T$ respectively, thus $Prob_{start}=p_{start}^{i},$$i\in{[1,\tau]}$ is the start probability vector for all tokens in $T$ and $Prob_{end}=p_{end}^{i},$$i\in{[1,\tau]}$ is the end probability vector for all tokens in $T$.

\begin{equation} 
\begin{split}
Prob_{start} = softmax(H^L*W_{start}+b_{start}) \\
Prob_{end} = softmax(H^L*W_{end}+b_{end})
\label{eq_prob_start}
\end{split}
\end{equation}

\subsubsection{Answerable Verification}
Motivated by Zhang's work, we introduce the same answerable verification step to determine whether an essay responds to a task requirement \cite{zhang2020retrospective}.

 We feed $h_1^L$ which is the representation vector of $[CLS]$ token encoded by ELECTRA into external front verification (E-FV) module. E-FV uses a fully connected layer followed by softmax operation to calculate classification logits $\hat{y}_i=(logit_{ans}, logit_{na})$ where $logit_{ans}$ is a scalar to indicate the answered logits and $logit_{na}$ is a scalar to indicate no-answer logits. We calculate the difference as the external verification score with Equation \ref{eq_score_ext}. 
 
Threshold-based answerable verification (TAV) takes start and end probabilities as input and outputs the no-answer score  $score_{diff}$ computed with Equation \ref{eq_score_has}, \ref{eq_score_null} and \ref{eq_score_diff}. $p_{start}^{1}$ and $p_{end}^{1}$ in Equation \ref{eq_score_null} represents the start and end probabilities of the $[CLS]$ token in $T$.

Rear verification (RV) combines $score_{diff}$ and $score_{ext}$ to obtain the final answerable score $score_{final}$ as shown in Equation \ref{eq_score_final}, where $\beta{1}$ and $\beta{2}$ are weights. MRC model predicts that question $q$ is answered by $E$ if $score_{final}>{\zeta}$, and not answered otherwise, where $\zeta$ is a hyper parameter.

\begin{subequations}
\begin{align}
score_{ext}=logit_{na}-logit_{ans} \label{eq_score_ext} \\
score_{has} = max(p_{start}^{k}+p_{end}^{l})\ k,l\in(1,\tau]\ and\  k \leq l \label{eq_score_has} \\
score_{null} = p_{start}^{1}+ p_{end}^{1}\label{eq_score_null} \\
score_{diff} = score_{null}-score_{has} \label{eq_score_diff} \\
score_{final} = \beta_{1}score_{diff}+\beta_{2}score_{ext}, \label{eq_score_final}
\end{align}
\end{subequations}

\subsection{Response Locating Module}
  In RL module, it takes start probabilities $Prob_{start}$, end probabilities $Prob_{end}$ and answerable score $score_{final}$ as input, and decides the start and end positions according to these inputs. A naive path to achieve this goal is that positions that obtains the highest start and end probabilities are chosen as start and end positions respectively. All tokens between these two positions are extracted as the student's response to the task requirement. If the start or the end position is less than $m+1$, in which case a span of question is marked, or their positions are contradictory, e.g., start position greater than end position, the module decides that the question is not responded. Finally, the framework outputs both the binary label indicating whether the student's essay does respond to the task requirement and the location of the responsive span if it is available.

\section{Experiments}
\label{sec:exp}

\subsection{Datasets}

\subsubsection{SQuAD 2.0}
SQuAD 2.0 is the most widely used benchmark in machine reading comprehension literature. It combines the first version of SQuAD with over 50,000 unanswerable questions written adversarially by crowd workers to look similar to answerable ones \cite{rajpurkar2016squad}. It contains 130,319 training examples from 442 Wikipedia articles and 11,873 development examples from 78 Wikipedia articles, where each example is made of a question and an article. This dataset requires that a model should not only answer the question when it is possible but also abstain from answering when there is no answer in the reading materials.

\subsubsection{SED}\label{subsection_writing_data}
This is a real-world student essay dataset that we collect from a third-party K-12 online learning platform. It consists of 9,450 examples in the training set and 3,357 examples in the test set, where each example contains an essay and a requirement question. There are 3,367 different essays and 593 different task requirement questions in the training set. In the test set, the number of essays and requirement questions are 1,655 and 185 respectively. In order to obtain labels, annotators need to 
firstly decide whether an essay does respond to the question and label it positive or negative accordingly. Secondly, for all positive essay examples, annotators need to mark the start and end positions of the span in the essay that responds to the question.

\begin{figure}[H]\centering
\includegraphics[scale=0.22]{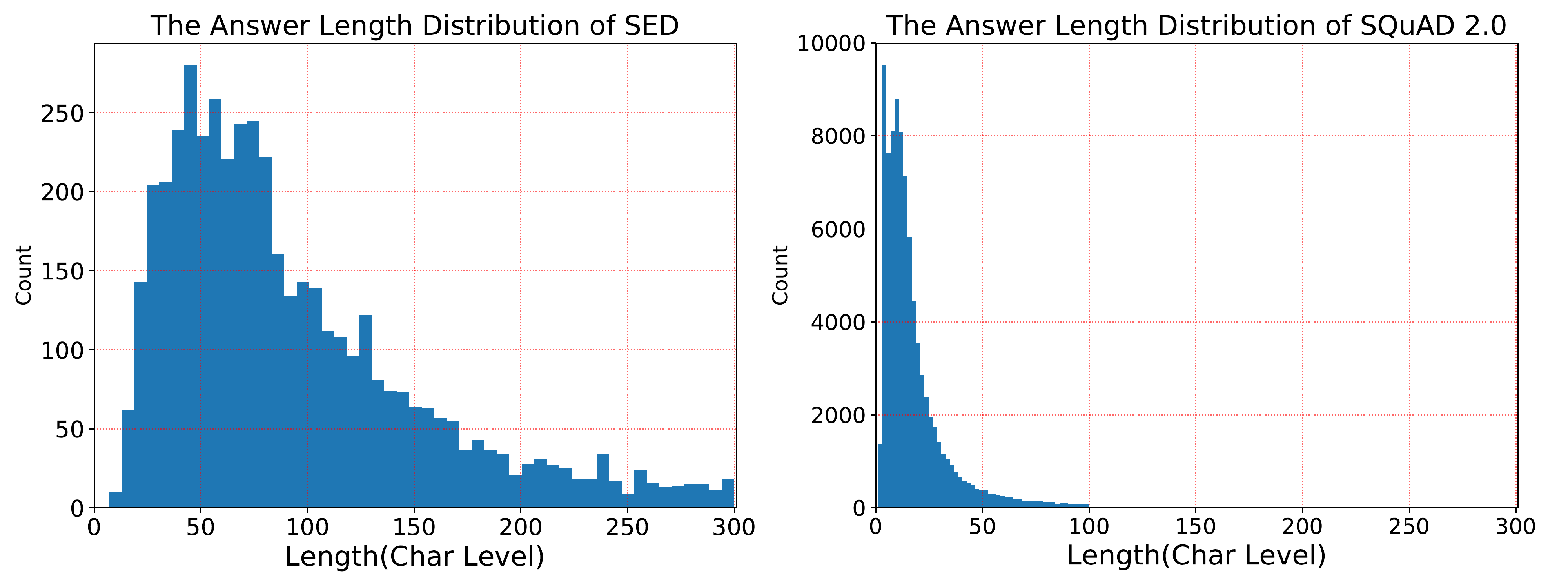}
\caption{Distributions of answer length (char level) in SED and SQuAD 2.0.}
\label{fig_distribution}
\end{figure}

Despite that SQuAD 2.0 and SED share similarities in terms of task and structure, there are many differences between them. First of all, SED is in the educational domain and SQuAD 2.0 is from Wikipedia. Secondly, answers in SED are much longer than answers in SQuAD 2.0. Fig \ref{fig_distribution} illustrates that most answers in SQuAD 2.0 are between 5 to 20 characters, while answers in SED are between 25 to 100 characters. The average length of answers in SQuAD 2.0 is 18.0 while the average length of answers in SED is 103.4. The last difference is that there are more grammatical errors in SED because essays in SED are written by second language learners. So a model that achieves the best performance on SQuAD 2.0 may not be directly deployed on educational scenarios.

\subsection{Experimental Setting}
In this section, we describe three sets of experiments as follows. 
\begin{itemize}
    \item Set 1. This set aims to prove that existing SOTA models on SQuAD 2.0 cannot be directly deployed on educational scenarios. In Set 1, all models are trained on SQuAD 2.0 but evaluated on the test set of SED. SAN was trained 50 epochs with learning rate $2e^{-3}$ on SQuAD 2.0 \cite{liu2017stochastic}. Pre-trained language models such as BERT, RoBERTa, ALBERT, and BART, were  acquired from hugging face\footnote{https://huggingface.co}. Our ELECTRA-based approach was trained 2 epochs with default parameters in this work \cite{zhang2020retrospective}. 
    \item Set 2. This set is to prove that MRC approaches are effective solutions to TRs writing evaluation when trained on the educational corpus. The training parameters of the models are consistent with those in Set 1. The difference is that models are all trained on SED.
    \item Set 3. This set explores how can we utilize SQuAD 2.0 and further improve model performance on SED. Following the idea that models pre-trained on massive data can be a good warm-up for subsequent finetuning, we first train MRC models on SQuAD 2.0 so as to acquire basic models, and then finetune them on SED for optimal performance.
\end{itemize}

%\vspace{-2em}
\begin{small}
\begin{equation}
\setlength{\abovedisplayskip}{-5mm}
\setlength{\belowdisplayskip}{-1mm}
 \begin{split}
    Accuracy = \frac{N_{correct}}{N_{total}} \\
    precision = \frac{ Num_{overlap}}{Num_{predict}} \\
    recall = \frac{Num_{overlap}}{Num_{gold}} \\
    F1 = \frac{2*{precision}*{recall}}{{precision}+{recall}}
\label{accuracy}
\end{split}
\end{equation}
\end{small}

%\vspace{-2em}
%\begin{equation}
% \begin{split}
%    Accuracy = N_{correct}/N_{total} \\
%    precision = Num_{overlap}/Num_{predict} \\
%    recall = Num_{overlap}/Num_{gold} \\
%    F1 = (2*{precision}*{recall})/({precision}+{recall})
%\label{accuracy}
%\end{split}
%\end{equation}

In all experiments, we use two evaluation indicators. One is Accuracy (Acc.) which measures the performance of the model on the binary classification task of predicting whether the essay answers the TR. Another is Answer Overlap F1 score (F1) which measures the performance of the model to predict the location of the answer span. Accuracy and F1 metrics can be calculated by Equations \ref{accuracy}.

%\begin{equation}
%	I(x_1,x_2) = \left\{
%\begin{array}{rcl}
%1       &      & {x_1  = x_2  }\\
%0     &      & {x_1 \neq x_2}\\
%\end{array} \right.
%	\label{indicator}
%\end{equation}
%\vspace{-1.0em}
%\begin{equation}
%    Accuracy = \frac{\sum\limits_{i=1}^{N}I(\hat{Y_i},Y_i)}{N}
%\label{accuracy}
%\end{equation}

%\vspace{-1em}

In Equation \ref{accuracy}, $N_{total}$ indicates the number of examples in the test set, and $N_{correct}$ is the number of examples that are correctly predicted by the framework. $Num_{overlap}$ represents the number of identical tokens in both the predicted span and the gold span. $Num_{predict}$ is the total number of tokens in predicted span and $Num_{gold}$ is the total number of tokens in the gold span.

\subsection{Results and Evaluation}
\subsubsection{Results of Set 1.} Table \ref{table_experiment_1} shows that existing SOTA models on SQuAD 2.0 are suffered a significant performance degradation on SED. All models in Table \ref{table_experiment_1} are well finetuned on SQuAD 2.0 and their F1 scores on SQuAD 2.0 dev set are all over 0.66. However, when evaluating them on SED test set, performances drop dramatically. For example, RoBERTa and our method achieve F1 score of 0.83 and 0.89 on SQuAD 2.0 dev set, but both drop to F1 score of 0.49 on SED test set.

\subsubsection{Result of Sets 2\&3.} Table \ref{table_experiment_2} shows results of Set 2 and Set 3. From the results of Set 2, we conclude that the MRC approaches can solve the TRs writing evaluation problem. Comparing with models trained on SQuAD 2.0 (Set 1), models trained on SED achieve significantly better results on SED test set. Our framework in Set 2 achieves the best F1 score of 0.84 and the best accuracy of 0.91, outperforms our framework in Set 1 by 23\% Accuracy and 35\% F1 score.

\FloatBarrier
\begin{table}[h]\centering

\caption{Performances of models trained on SQuAD 2.0}
\label{table_experiment_1}
\scalebox{0.9}{
\setlength{\tabcolsep}{2.5mm}{
\begin{tabular}{llll|ll}
\hline
\multirow{2}{*}{Training Dataset} & \multirow{2}{*}{Methods} & \multicolumn{2}{l}{SQuAD 2.0 dev} & \multicolumn{2}{|l}{SED test} \\
                                  &                         & ACC.             & F1             & ACC.           & F1          \\
\hline \\[-1em]
\multirow{6}{*}{SQuAD 2.0 (Set 1)\ \ \ \ \ \ } & SAN                   &      0.70          &      0.66         &       0.57          &     0.31          \\
& BERT                  &      0.78          &      0.73         &       0.65          &     0.37          \\
& ALBERT                &      0.85          &      0.81         &       0.58          &     0.37          \\
& RoBERTa               &      0.86          &      0.83         &       0.69          &     0.49          \\
& BART                  &      0.87          &      0.81         &       0.62          &     0.42          \\
& Ours &      0.92          &      0.89         &       0.68          &     0.49        
\\ \\[-1em]
\hline                                                     
\end{tabular}
}
}
\end{table}
\FloatBarrier

%\vspace{-1em}

If we compare results in Set 2 and Set 3, we find that optimal performance can be obtained by firstly training models on SQuAD 2.0 and then finetuning on SED. Specifically, F1 score of SAN increases by 11\%, and F1 score of BERT increases by 8\%. Similarly, the accuracy also increases significantly in Set 3.

\FloatBarrier
\begin{table}[h]
\centering
\caption{Performances of models trained on SED and SQuAD 2.0\&SED}
%$\dagger$ represents sentence level approach.
\label{table_experiment_2}
\scalebox{0.9}{
\setlength{\tabcolsep}{2.5mm}{
\begin{tabular}{@{}clcc@{}}
\hline
\multicolumn{1}{l}{\multirow{2}{*}{Training Dataset}} & \multirow{2}{*}{Methods} & \multicolumn{2}{c}{SED Test} \\
\multicolumn{1}{l}{}                                  &                         & Acc.           & F1          \\
\hline \\[-1em]
\multirow{8}{*}{SED (Set 2)}             
%& GDBT$\dagger$            &   0.62       &  0.42  \\
%                                     & RoBERTa$\dagger$        &   0.73       &  0.59  \\
                                     & SAN                   &   0.67       &  0.58 \\
                                     & BERT                  &   0.79       &  0.68 \\
                                     & ALBERT                &   0.84       &  0.77  \\
                                     & RoBERTa               &   0.81       &  0.71  \\
                                     & BART                  &   0.82       &  0.73  \\
                                     & Ours &   0.91       &  0.84  \\
                                     \hline \\[-1em]
 \multirow{6}{*}{SQuAD 2.0\&SED (Set 3)}      & SAN                   &   0.79 (+0.12)        &  0.69 (+0.11)   \\
                                     & BERT                  &   0.84 (+0.05)        &  0.76 (+0.08) \\
                                     & ALBERT                &   0.86 (+0.02)      &  0.80 (+0.03) \\
                                     & RoBERTa              &   0.88 (+0.07)       &  0.80 (+0.09) \\
                                     & BART                  &   0.89 (+0.07)      &  0.82 (+0.09) \\
                                     & Ours &   \textbf{0.93 (+0.02) }       &  \textbf{0.85 (+0.01)}  \\ \\[-1em]
\hline
\end{tabular}
}
}
\end{table}
\FloatBarrier

Comparing with Set 2, the accuracy of BART and our framework increase by 7\% and 2\% respectively. Furthermore, our approach achieves the best performance in each of the three sets of experiments, and outperforms a variety of SOTA approaches.

\section{Conclusion}
\label{sec:conclusion}
%Detailed feedback is extremely important for students to improve their writing skills. Existing AES systems focus on giving a holistic score on students' essays but can not provide details to support the grading of task requirement writing. 

In this paper, we proposed a MRC based approach which cannot only detect if an essay responds to a requirement question but find where the essay answers the question. From our experiments and analysis, we demonstrate that SQUAD 2.0 is very different from our educational dataset, so existing SOTA models on SQuAD 2.0 cannot be directly deployed on educational scenarios. Instead, we propose to firstly train a basic model on SQuAD 2.0 and then finetune the basic model on educational data for optimal performance. We believe this proposed framework is able to help automatic essay scoring systems provide detailed grading results, thereby helping students improve their writing skills.

%We prove the effectiveness of MRC approaches. We have also explored a way to effectively use SQuAD to improve the model's performance on SED. And we also explored a variety of prevalence MRC models including SAN, RoBERTa and Retro-Reader(ELECTRA). Retro-Reader(ELECTRA) gets the best performance on SED among other models.

%Our future work will focus on improving the Question Rewriting Module, enhancing its revising ability to generate questions more friendly to the MRC model. In practical production environment, we discover that the phraseology of question has an influence on the performance of our system. Of course we could preprocess these question factitiously, but it is better to use a rephrase model to do this onerous and boring work automatically considering there has been a lot of researches and progresses in the field of rephrase.

\section*{Acknowledgment}
\label{sec:acknowledgement}
This work was supported in part by National Key R\&D Program of China, under Grant No. 2020AAA0104500 and in part by Beijing Nova Program (Z201100006820068) from Beijing Municipal Science \& Technology Commission.

\bibliographystyle{splncs04.bst}
\bibliography{aied2021}
\end{document}